\documentclass{article} 
\usepackage{arxiv_preprint,times}

\usepackage{amsmath,amsfonts,bm}

\def\eqref#1{equation~\ref{#1}}

\def\1{\bm{1}}

\DeclareMathAlphabet{\mathsfit}{\encodingdefault}{\sfdefault}{m}{sl}
\SetMathAlphabet{\mathsfit}{bold}{\encodingdefault}{\sfdefault}{bx}{n}

\usepackage{hyperref}
\usepackage{url}
\usepackage{booktabs}
\usepackage[table]{xcolor}
\usepackage{graphicx}
\usepackage{bbm}
\usepackage{multirow}
\usepackage{xcolor}
\usepackage{makecell}
\usepackage{amssymb}
\usepackage{wrapfig}

\newcommand{\fpvalue}[1]{\textcolor{gray!75}{#1}}
\newcommand{\venue}[1]{%
  {\fontsize{5.5}{6.5}\selectfont
  \textcolor{gray!70}{\textnormal{(#1)}}}%
}
\newcommand{\rowtag}[1]{%
  {\fontsize{6.5}{6.5}\selectfont
  \textcolor{gray!70}{\textnormal{(#1)}}}%
}

\title{P4Q: Co-designing Token Pruning and Quantization for Vision-Language Model Acceleration}

\author{
Haizhao Jing$^{1}$, Zhenhao Shang$^{1}$, Haokui Zhang$^{1}$, Rong Xiao$^{2}$, Peng Wang$^{1}$ \\[0.5em]
\normalfont $^{1}$Northwest Polytechnical University \\
$^{2}$Intellifusion
}

\begin{document}

\maketitle

\begin{abstract}
Vision language models have achieved strong performance across a wide range of multimodal applications, yet their substantial computational and memory costs hinder efficient deployment. Visual token pruning and post-training quantization reduce inference overhead along two complementary dimensions, namely sequence length and numerical precision. Existing workflows typically optimize these techniques independently or apply them sequentially. Their distinct optimization objectives leave critical interactions unaddressed and constrain the achievable compression performance. 
We revisit these designs and present P4Q, a practical co-design framework that jointly optimizes visual token pruning and low-bit quantization for efficient VLM inference. First, P4Q introduces a quantization-aware visual token selection strategy before the LLM. It applies fake quantization to copies of the features produced by the projector and selects visual tokens using statistics computed from these fake-quantized features, thereby conditioning the selector’s feature-based decisions on simulated low-bit perturbations. Second, P4Q introduces a pruning-aware quantization calibration strategy. It uses the same selection strategy as pruning to calibrate the quantized model on the retained-token distribution, thereby aligning the calibration process with the pruned execution path used during deployment. By coupling these two components, P4Q achieves substantial inference speedups while maintaining comparable task performance, resulting in a better efficiency-accuracy trade-off than independently optimized pipelines. For instance, on LLaVA-NeXT, P4Q achieves an average end-to-end inference speedup of 2.8$\times$ across eight distinct test sets, while retaining higher accuracy than prior compression and quantization methods.



\end{abstract}

\section{Introduction}
\label{sec:introduction}

Modern VLMs commonly encode each image into hundreds or thousands of visual tokens, although these representations exhibit substantial redundancy \citep{chen2024fastv,alvar2025divprune,shang2025llavaprumerge}. This inefficiency has motivated two complementary compression paradigms: post-training quantization (PTQ) and visual token pruning. PTQ reduces the numerical precision of model weights and activations, lowering memory and computation costs through calibrated low-bit representations \citep{sun2025flatquant,li2025mbq,wang2024qvlm}. Visual token pruning instead shortens the multimodal sequence by removing redundant or less informative image tokens before or within the language model. Applying both should reduce the cost of each operation and the number of tokens on which it operates. The difficulty is that the two tools also change the conditions under which the other was designed.

Current visual token pruning methods typically identify and remove redundant or weakly informative visual tokens before or within the language model to reduce subsequent computation \citep{chen2024fastv,huang2024ivtp,alvar2025divprune}. However, recent studies show that attention-based importance estimates can be unstable, poorly aligned with actual token utility, and spatially biased \citep{wen2025tokenpruning,zhang2025vispruner,endo2025feather}. This weakness becomes particularly consequential near the top-$K$ boundary: when two tokens have a small score margin, a perturbation exceeding that margin can reverse their order and discontinuously change the retained set. We further observe that low-bit quantization introduces an additional source of such perturbations, as errors accumulated in hidden states and linear transformations propagate to the attention scores used for selection. Consequently, under low-bit execution, attention-based importance estimation inside the language model becomes more unstable, making the retained visual-token set increasingly sensitive to quantization-induced perturbations.

\begin{figure*}[t]
    \centering
    \includegraphics[width=0.85\columnwidth]{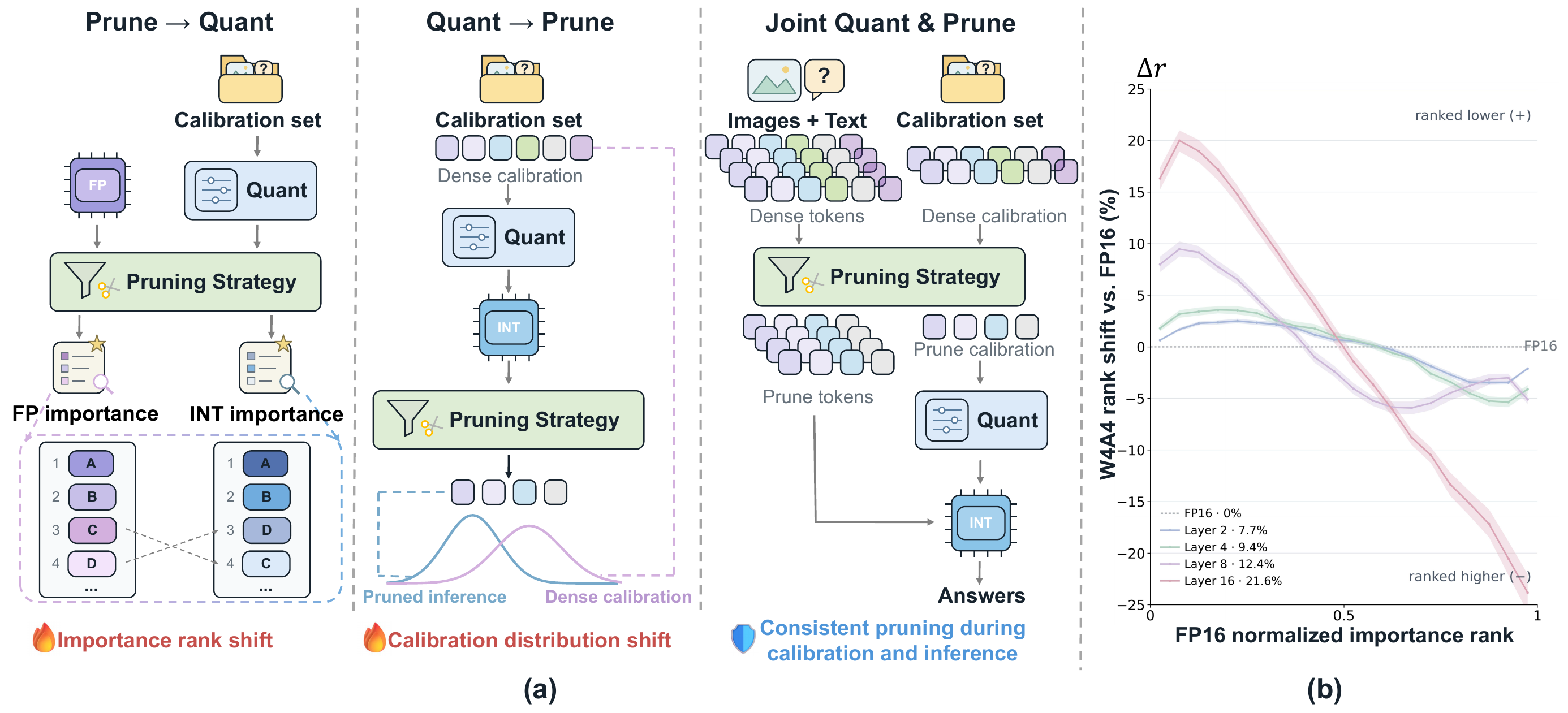}
    \caption{\textbf{Comparison of sequential and joint pipelines for low-bit calibration and visual-token pruning, together with a visualization of the visual-token importance ranking shifts induced by low-bit quantization across LLM layers.}
    \textbf{(a)} Prune-then-quantize suffers from quantization-induced importance-rank shifts, whereas quantize-then-prune creates a calibration-inference distribution shift. P4Q jointly prunes inference and calibration sequences using the same selector and token budget, matching their retained-token support and sequence length.
    \textbf{(b)} W4A4 induces layer-dependent visual token rank shifts relative to FP16 across representative in-LLM pruning layers in LLaVA-NeXT-7B.}
    \label{fig:p4q_motivation}
\end{figure*}

A parallel mismatch arises during quantization calibration. Calibration-based PTQ fits low-bit parameters to the activation distributions induced by the calibration data. Specifically, calibration samples generate layer-wise activations from which quantization scales, clipping ranges, and transformation parameters are estimated. These parameters are therefore specialized to the activation regime observed during calibration. Prior studies show that calibration-data selection can substantially affect the downstream performance of compressed LLMs \citep{williams-aletras-2024-impact}. A mismatch between calibration and inference sequence lengths can further degrade quantized-model accuracy \citep{lee-etal-2023-enhancing-computation}. VLM-specific studies further demonstrate the importance of calibration design at low bit widths \citep{wang2024qvlm,li2025mbq}. Q-VLM optimizes layer-wise rounding while accounting for cross-layer dependencies, whereas MBQ employs modality-balanced reconstruction to address unequal vision language sensitivity. This dependence creates a systematic mismatch when a densely calibrated VLM is subsequently combined with visual-token pruning. During deployment, each sample presents the low-bit language model with a shorter, sample-dependent sequence than that used during calibration. Pruning alters the retained visual-token support, attention context, sequence length, and downstream activation statistics. Dense calibration therefore optimizes quantization parameters for a computational graph and activation distribution that differ from those encountered during pruned inference.

Taken together, independently optimizing or sequentially applying visual token pruning and quantization creates two coupled mismatches. Quantization perturbations can distort visual-token selection, while dense-sequence calibration fits the quantizer to an execution path different from that encountered during pruned inference.

To study these interactions, we introduce P4Q, a practical co-design framework that jointly optimizes visual token pruning and low-bit quantization for efficient VLM inference. First, P4Q introduces a quantization-aware visual token selection strategy before the LLM. The selector determines token indices from fake-quantized copies of projector features without using quantized decoder states for scoring. Second, P4Q introduces a pruning-aware quantization calibration strategy. For each calibration sample, it uses the inference-time selector and token budget to construct the retained-token sequence on which the quantizer is fitted. By coupling these two components, P4Q achieves substantial inference speedups while preserving strong task performance, providing a better efficiency-accuracy trade-off than independently optimized pipelines. Our main contributions are summarized as follows.

\begin{itemize}
    \item We rethink visual token pruning and quantization as a coupled compression problem and introduce P4Q, a practical co-design framework for efficient VLM inference. P4Q employs a quantization-aware pre-LLM selector that uses fake-quantized copies of projector features in its feature-based scoring stages, allowing token selection to account for simulated low-bit perturbations. It further performs pruning-aware PTQ calibration on sequences produced by the same selector and token budget used at inference, exposing the quantizer to the resulting retained-token support and sequence length.

    \item We validate P4Q across eight multimodal benchmarks under W4A4 quantization and an 88.9\% visual token pruning ratio. On LLaVA-NeXT-7B, P4Q achieves an average $2.8\times$ end-to-end inference speedup while reducing peak GPU memory usage to 51.3\% of the vanilla baseline. Under the same setting, P4Q achieves an average $2.49\times$ end-to-end speedup on LLaVA-1.5-7B and reduces peak GPU memory usage by 58\% relative to the vanilla baseline.
    



\end{itemize}

\section{Related Work}
\label{sec:related_work}

\paragraph{Visual Token Pruning for VLMs.} VLMs typically encode an image as a long sequence of patch tokens, many of which are redundant for a given input and task. A prominent family of pruning methods determines which tokens to retain inside the language model, using cross-modal attention or intermediate representations as task-aware signals. FastV, for example, ranks visual tokens using text-to-vision attention from a shallow decoder layer and removes weakly attended tokens from subsequent computation \citep{chen2024fastv}. Although such signals reflect explicit vision language interaction, they are also conditioned on the decoder execution path and the layer at which they are measured. This conditionality may make the retained set sensitive to the decoder-side scoring process. \cite{wen2025tokenpruning} systematically show that attention-based importance does not always identify useful visual tokens reliably and that carefully designed pruning methods can even underperform random selection. FEATHER further demonstrates that early decoder attention can inherit RoPE-induced positional bias, producing spatially skewed retained sets and pronounced degradation on localization-intensive tasks \citep{endo2025feather}. Together, these findings suggest that decoder-side importance is not an intrinsic property of a visual token, but a conditional estimate shaped by attention geometry, pruning depth, input characteristics, and model state. 

Against this backdrop, a growing line of work performs visual-token selection outside the language model, deriving retention signals from vision-side representations or lightweight pre-LLM interactions. LLaVA-PruMerge uses sparse class-to-patch attention from the vision encoder to identify visual anchors, then clusters and merges the remaining tokens to supplement the selected content before it reaches the LLM \citep{shang2025llavaprumerge}. DivPrune removes reliance on decoder attention by formulating pre-LLM selection as a max--min diversity problem, thereby preserving broad coverage of the visual feature space \citep{alvar2025divprune}. CRISP restores instruction dependence before decoder execution by grounding question nouns in projected visual features and completing the retained set with diversity-based contextual tokens \citep{crisp}. Collectively, these methods construct retained sets from vision-side saliency, feature coverage, or lightweight task cues rather than relying on a single decoder-side importance signal.

\paragraph{Post-Training Quantization for VLMs.} PTQ enables low-bit deployment by reducing the precision of model weights and/or activations without full-model retraining, thereby lowering storage and memory-traffic costs. Unlike text-only LLMs, however, VLMs couple a vision encoder and projector with a language backbone, while visual and textual tokens exhibit different activation statistics and sensitivities to quantization error. Q-VLM addresses cross-layer error dependencies by using activation entropy to partition correlated blocks and search their rounding strategies jointly \citep{wang2024qvlm}. MBQ instead accounts for modality-specific sensitivity by reweighting the reconstruction loss for visual and textual tokens \citep{li2025mbq}. These methods establish that VLM quantization requires calibration objectives adapted to multimodal inference rather than homogeneous, layer-isolated reconstruction.

Recent work further refines the granularity and execution fidelity of calibration. QIG uses quantization-aware integrated gradients to estimate token-level sensitivity and capture inter- and intra-modality interactions after visual and textual representations have been mixed \citep{xiang2026qig}. VLM-PTQ corrects residual-induced target shifts and combines modality-specific Hessian estimates to distinguish channel importance across visual and textual tokens \citep{deng2026vlmptq}. TLQ constructs a token-importance-aware calibration set and exposes each layer-wise calibration stage to the preceding quantized path, bringing calibration closer to the actual low-bit forward pass \citep{shang2026tlq}. Collectively, these methods show that calibration data, error weighting, optimization targets, and forward-path simulation are integral components of VLM quantization design.

\paragraph{Joint Quantization and Visual Token Pruning.} Quantization and visual token pruning reduce complementary inference costs by lowering numerical precision and shortening the sequence processed by the language backbone, respectively. Joint methods therefore coordinate token selection with the low-bit regime used at deployment. In the vision language action setting, SQAP-VLA applies quantization-aware pruning criteria to an aggressively quantized model and adapts the quantizer to improve pruning effectiveness \citep{fang2025sqapvla}. For multimodal language models, QAPruner scores tokens using semantic relevance, simulated group-wise quantization error, and activation-outlier intensity, thereby accounting for both task relevance and low-bit numerical stability \citep{wang2026qapruner}. QUOTA instead aligns calibration with pruned execution by converting low-bit calibration signals into an offline layer-wise token-allocation schedule and evaluating token importance under deployed W4A4 operators and a quantized KV cache \citep{li2026quota}. Together, these methods couple compression through quantizer adaptation, quantization-aware token scoring, or calibration-derived pruning schedules, marking a shift from independently applied compressors toward deployment-aware joint design.

\section{Method}
\label{sec:method}
In this section, we first motivate our method by examining how the serial composition of visual-token pruning and PTQ creates token-ranking and calibration-distribution mismatches. We then introduce P4Q, detailing how to select complementary visual evidence before the low-bit language backbone and calibrate the quantized model on sequences pruned using the inference-time selector and token budget.

\begin{figure*}[t]
    \centering
    \includegraphics[width=0.75\linewidth]{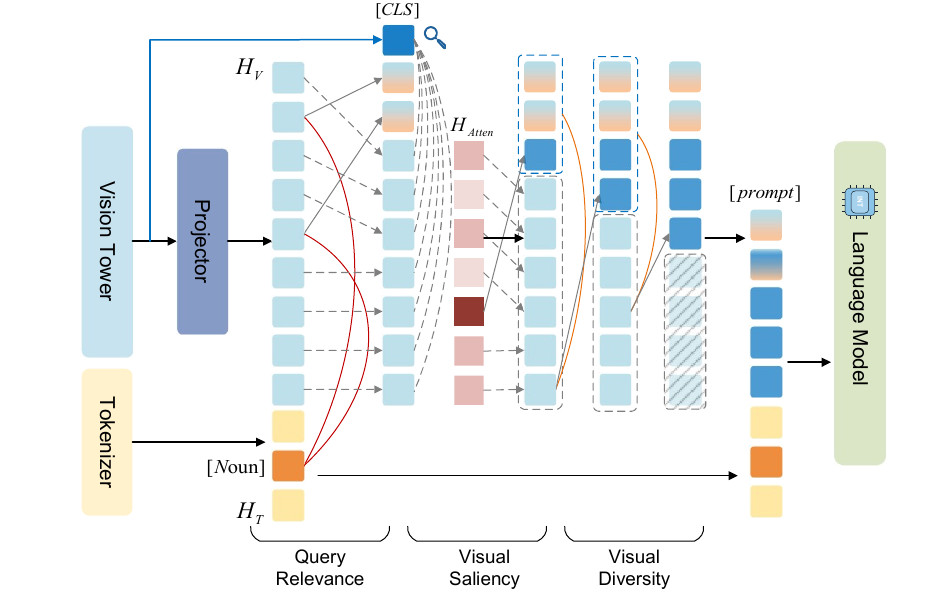}
    \caption{Overview of quantization-aware pruning in P4Q. It first retains query-relevant tokens to capture task semantics, then selects visually salient tokens based on vision-tower CLS-to-patch attention, and finally applies max-min diversity sampling to reduce redundancy and broaden visual coverage. The query relevance and  visual diversity stages use fake-quantized feature copies, while visual saliency uses vision tower attention.
    }
    \label{fig:p4q_pipeline}
\end{figure*}

\subsection{Motivation}
\label{sec:motivation}


\paragraph{Low-bit visual token ranking shift.}
Existing in-LLM pruning methods commonly estimate token importance from language-model attention or hidden activations, rank the visual tokens accordingly, and retain the Top-$K$ tokens. Let $\mathbf{s}_{\ell}^{\mathrm{FP}}$ denote the importance scores obtained from the FP16 states at layer $\ell$. When the same estimator is applied under low-bit execution, it instead receives quantized hidden states and attention statistics, yielding
\begin{equation}
\mathbf{s}_{\ell}^{\mathrm{INT}}
=
\Psi\!\left(H_{\ell}^{\mathrm{INT}},A_{\ell}^{\mathrm{INT}}\right)
=
\Psi\!\left(H_{\ell}^{\mathrm{FP}},A_{\ell}^{\mathrm{FP}}\right)
+
\boldsymbol{\delta}_{\ell},
\qquad
\mathcal{S}_{\ell,K}^{\mathrm{INT}}
=
\operatorname{TopK}\!\left(\mathbf{s}_{\ell}^{\mathrm{FP}}+\boldsymbol{\delta}_{\ell},K\right),
\label{eq:int_token_selection}
\end{equation}
where $\Psi$ is the importance estimator, and $H_{\ell}$ and $A_{\ell}$ denote the hidden states and attention statistics available at layer $\ell$, respectively. The perturbation $\boldsymbol{\delta}_{\ell}$ captures the token-dependent changes in importance scores induced by low-bit computation. The corresponding FP16 selection is $\mathcal{S}_{\ell,K}^{\mathrm{FP}}=\operatorname{TopK}(\mathbf{s}_{\ell}^{\mathrm{FP}},K)$. Because Top-$K$ selection depends on relative ordering rather than absolute score magnitude, even moderate perturbations can reverse the order of tokens with similar FP16 scores, particularly near the selection boundary. Consequently, $\mathcal{S}_{\ell,K}^{\mathrm{INT}}$ may differ from $\mathcal{S}_{\ell,K}^{\mathrm{FP}}$, producing a quantization-induced shift in the retained visual-token support.


To visualize quantization-induced ranking drift, we align each visual token between the FP16 and W4A4 runs using its original packed index and define its signed normalized rank displacement as:
\begin{equation}
\Delta r_{\ell}^{\mathrm{INT}\leftarrow\mathrm{FP16}}
=
r_{\ell}^{\mathrm{INT}}
-
r_{\ell}^{\mathrm{FP16}}
=
\frac{
\operatorname{rank}_{\downarrow}
\!\left(s_{\ell}^{\mathrm{INT}}\right)
-
\operatorname{rank}_{\downarrow}
\!\left(s_{\ell}^{\mathrm{FP16}}\right)
}{N-1},
\label{eq:visual_token_rank_shift}
\end{equation}
where $s_{\ell}$ denotes the importance score of a visual token at layer $\ell$, and $\operatorname{rank}_{\downarrow}$ assigns rank one to the largest score. Accordingly, $\Delta r>0$ indicates that a token moves toward a less important position under W4A4, whereas $\Delta r<0$ indicates a move toward a more important position. We evaluate this displacement on 128 fixed samples using identical dense inputs, attention masks, positional indices, and packed token indices. In Fig.~\ref{fig:p4q_motivation} (b), tokens are grouped by their same-layer FP16 normalized importance rank, from the most important ($0$) to the least important ($1$), and each curve reports the sample-balanced mean displacement at a representative in-LLM pruning layer of LLaVA-NeXT-7B. W4A4 quantization induces substantial rank rearrangement at representative in-LLM pruning layers, indicating that token importance rankings derived under FP16 precision are not necessarily preserved within a low-bit language backbone. More details are provided in Appendix~\ref{sec:appendix_rank_shift}.

\paragraph{Quantization-aware selection requires low-bit statistics.} Recent studies have begun to show that pruning strategies intended for low-bit deployment should account for the numerical regime in which their selection statistics are produced. In particular, quantization can distort the attention or feature statistics used for token ranking, making selection criteria derived exclusively from full-precision states unreliable when directly applied to quantized models~\citep{fang2025sqapvla}. Quantization-aware pruning methods consequently evaluate token importance using quantized-model statistics or explicitly incorporate simulated quantization errors into the selection criterion~\citep{wang2026qapruner}. These observations motivate us to test whether pruning under low-bit inference benefits from token scoring that accounts for simulated perturbations of the target quantization configuration.

\paragraph{Pruning-induced calibration mismatch.}
Calibration-based post-training quantization is inherently dependent on the activation distribution induced by the calibration data. Calibration samples are forwarded through the model to generate layer-wise activations, from which the quantization scales, clipping ranges, and transformation parameters are estimated. Consequently, these parameters are optimized for the activation regime observed during calibration rather than being independent of the calibration inputs. Prior studies have shown that the choice of calibration data can cause substantial variation in the downstream performance of compressed LLMs~\citep{williams-aletras-2024-impact}. In particular, a mismatch between the sequence lengths used for calibration and inference can degrade the accuracy of the quantized model, highlighting the importance of aligning the calibration sequence length with the deployment-time inference sequence length~\citep{lee-etal-2023-enhancing-computation}. 

Specifically, this dependence can be formalized through the calibration objective. Let $H_{\ell,n}^{\mathrm{dense}}$ denote the input activation of language-model block $\ell$ for calibration sample $n$ when all visual tokens are retained. Under conventional dense calibration, the quantization parameters are obtained by minimizing the reconstruction error between the full-precision block and its quantized counterpart:
\begin{equation}
    \theta_{\ell,\mathrm{dense}}^{*}
    =
    \arg\min_{\theta_{\ell}}
    \frac{1}{B}
    \sum_{n=1}^{B}
    \operatorname{MSE}
    \left(
        f_{\ell}^{\mathrm{INT}}
        \left(H_{\ell,n}^{\mathrm{dense}}\right),
        f_{\ell}^{\mathrm{FP}}
        \left(H_{\ell,n}^{\mathrm{dense}}\right)
    \right),
    \label{eq:dense_calibration_objective}
\end{equation}
where $B$ is the number of calibration samples, $f_{\ell}^{\mathrm{FP}}$ and $f_{\ell}^{\mathrm{INT}}$ denote the full-precision block and its low-bit counterpart, respectively, and $\theta_{\ell}$ contains the quantization parameters optimized during calibration. Therefore, $\theta_{\ell,\mathrm{dense}}^{*}$ is fitted to the activation distribution induced by dense visual sequences. After visual token pruning, however, the same block receives $H_{\ell,n}^{\mathrm{pruned}}$, whose sequence length and visual-token support are determined by the retained tokens, thereby altering the attention support, the activation distributions observed by subsequent blocks, and consequently the quantization parameters $\theta_{\ell}$ obtained during calibration.
These findings from prior studies, together with the preceding theoretical analysis, indicate that effective PTQ calibration should align the calibration-data distribution, sequence structure, and resulting activation ranges with those encountered during quantized inference.

\subsection{Quantization-Aware Pre-LLM Selection}
\label{sec:pre_llm_pruning}
Motivated by the low-bit visual-token ranking shift and the need for quantization-consistent selection statistics identified in Section~\ref{sec:motivation}, P4Q performs quantization-aware visual token selection before the language backbone. This placement avoids relying on progressively perturbed attention values or hidden states produced inside the quantized decoder to determine the retained token set. P4Q applies selector-only fake quantization to the projector features used by its feature-based scoring stages. This exposes those selection decisions to a simulated low-bit perturbation. The fake-quantized projector features are used exclusively to determine the retained token indices, whereas the corresponding original FP16 representations are forwarded to the language backbone for subsequent inference. As illustrated in Fig.~\ref{fig:p4q_pipeline}, P4Q further integrates three complementary priors to enrich the visual evidence preserved by the selected tokens: query-relevant information, intrinsically salient visual content, and complementary visual information that improves the diversity of the retained set.

Specifically, P4Q partitions the total retention budget among three sequential signals using the budget ratios $\alpha$, $\beta$, and $\gamma$.
The overall selection is expressed as:

\begin{equation}
\mathcal{S}_{\mathrm{keep}}
=\underbrace{\alpha\operatorname{TopK}_{i\in\mathcal{S}_{\mathrm{v}}}
(\operatorname{CosSim}
\left(
\widetilde{\mathbf{q}}_j,
\widetilde{\mathbf{v}}_i
\right))}_{\text{Query Relevance}}
+
\underbrace{\beta\operatorname{TopK}_{i\in\mathcal{S}_{\mathrm{v'}}}
\left(A_i^{\mathrm{cls}}\right)}_{\text{Visual Saliency}}
+
\underbrace{\gamma\mathrm{Div}
\left(\operatorname{CosSim}
\left(\widetilde{\mathbf{v}}_i,\widetilde{\mathbf{v}}_d\right)
\right)}_{\text{Visual Diversity}}.
\label{eq:p4q_progressive_selection}
\end{equation}

Here, $\mathcal{S}_{\mathrm{v}}$ denotes the complete set of visual token indices, $\mathbf{q}_j$ is the mean subword embedding of the $j$-th noun or proper noun extracted from the question, and $\widetilde{\mathbf{v}}_i$ and  $\widetilde{\mathbf{q}}_j$ are fake-quantized copies of the projector feature. The fake-quantized features are used only by the selector, whereas the corresponding original FP16 features are forwarded to the language backbone after selection.

The three signals are applied sequentially and serve complementary purposes. First, the \textbf{Query Relevance} stage independently ranks the visual tokens for each query concept according to the cosine similarity between $\widetilde{\mathbf{q}}_j$ and $\widetilde{\mathbf{v}}_i$. For every concept embedding $\mathbf{q}_j$, P4Q constructs a concept-specific ranking of the candidate visual tokens and retains the highest-scoring tokens within the allocated quota. This concept-wise selection preserves localized visual evidence associated with multiple entities or attributes and prevents a single dominant concept from occupying the entire semantic budget. However, the resulting signal captures only visual evidence directly aligned with the explicit noun concepts in the query and may overlook salient objects or scene-level context that is not explicitly mentioned.

P4Q therefore performs \textbf{Visual Saliency} selection over the remaining visual-token sequence $\mathcal{S}_{\mathrm{v'}}$, which contains the tokens not retained by the preceding query-guided stage. Specifically, P4Q extracts the CLS-to-patch attention weights from the final multi-head self-attention layer of the vision tower. In this attention operation, the final-layer [CLS] token serves as the query, while the patch tokens indexed by $\mathcal{S}_{\mathrm{v'}}$ serve as the candidate visual keys. The attention weights are then averaged across heads to obtain the saliency score $A_i^{\mathrm{cls}}$ for each remaining visual token $i$. The tokens with the highest saliency scores are retained within the budget, supplementing the query-relevant evidence with intrinsically salient objects and global scene information.

Although the first two stages preserve query relevant and visually salient evidence, the selected tokens may remain concentrated in a small number of semantically or spatially correlated regions. The remaining budget is therefore allocated to \textbf{Visual Diversity} completion. Specifically, for each token in the remaining candidate pool, the maximum cosine similarity to any tokens in the current retained set is computed. The candidate with the lowest maximum similarity is retained, as it is the least redundant with the current selection. After each selection, the current retained set is updated and the similarity scores of the remaining candidates are recomputed. This max-min procedure is repeated until the allocated budget is exhausted, progressively extending the retained visual coverage to complementary objects, background regions, and spatial context.

We set the budget ratios to $\alpha=0.3$, $\beta=0.2$, and $\gamma=0.5$, allocating 30\%, 20\%, and 50\% of the retained-token budget to Query Relevance, Visual Saliency, and Visual Diversity, respectively. Performance remains stable under reasonable variations of these ratios, indicating that P4Q is not sensitive to their exact values. Detailed ablation results are provided in Appendix~\ref{sec:appendix_pruning}.

\subsection{Pruning-Aware Quantization Calibration}
\label{sec:purning-aware ptq}

Post-training quantization estimates its transformation and clipping parameters from the activation distributions observed during calibration. Dense calibration therefore produces parameters adapted to sequences containing all visual tokens, whereas the deployed model operates on shorter and sample-dependent sequences after pruning. Visual-token removal changes the token support and sequence length processed by the language model, which in turn modifies attention interactions and the activation distributions of subsequent blocks. To reduce this mismatch, we apply the deployment-time visual token selector before collecting calibration activations.

Specifically, we construct a calibration set of 128 samples using a fixed random seed. For each sample, the fixed P4Q selector receives only the image and question and generates the pruned multimodal sequence using exactly the same token budget and selection configuration as during inference. We then perform block-wise quantization calibration on the resulting pruned activations. For language-model block $\ell$, the calibration objective is:
\begin{equation}
\theta_{\ell,\mathrm{keep}}^{*}
=
\arg\min_{\theta_{\ell}}
\frac{1}{B}
\sum_{n=1}^{B}
\frac{1}{m_n^{\mathrm{keep}}d}
\left\|
f_{\ell,\theta_{\ell}}^{\mathrm{W4A4}}
\left(H_{\ell,n}^{\mathrm{FP,keep}}\right)
-
f_{\ell}^{\mathrm{FP}}
\left(H_{\ell,n}^{\mathrm{FP,keep}}\right)
\right\|_{F}^{2}.
\label{eq:p4q_pruning_aware_calibration}
\end{equation}
Here, $B=128$ is the number of calibration samples, and $n$ and $\ell$ index the sample and language-model block, respectively. $H_{\ell,n}^{\mathrm{FP,keep}}$ denotes the input to block $\ell$ obtained by propagating the P4Q-retained sequence through the preceding full-precision blocks, where $\mathrm{keep}$ distinguishes this sequence from its dense counterpart. $m_n^{\mathrm{keep}}$ is the corresponding multimodal sequence length, and $d$ is the hidden dimension. The functions $f_{\ell}^{\mathrm{FP}}$ and $f_{\ell,\theta_{\ell}}^{\mathrm{W4A4}}$ represent the full-precision block and its W4A4 fake-quantized counterpart. Their outputs are reconstructed from the same pruned input by optimizing the diagonal channel scales and the weight- and activation-clipping parameters in $\theta_{\ell}$. Calibration proceeds block by block, with the full-precision output of block $\ell$ serving as the input to block $\ell+1$, thereby avoiding the accumulation of upstream quantization errors. $\theta_{\ell,\mathrm{keep}}^{*}$ denotes the resulting optimal parameters for the P4Q-pruned calibration path. Further details are provided in Appendix~\ref{sec:appendix_calibration_implementation}.

\section{Experiments}
\label{sec:experiments}

\subsection{Experimental Setup}
\label{sec:experimental setup}
\textbf{Benchmarks and Baselines.}
We evaluate P4Q on eight established image-understanding benchmarks: MMMU~\citep{yue2024mmmu}, VizWiz~\citep{gurari2018vizwiz}, ScienceQA-TEST (SQA)~\citep{lu2022learnsqa}, MME~\citep{fu2025mme}, GQA~\citep{hudson2019gqa}, MMBench-CN (MMB$^{\mathrm{CN}}$)\citep{liu2024mmbench}, POPE~\citep{li2023evaluatingpope}, and SEEDBench-IMG (SEED)~\citep{li2023seed}. We compare against five representative baselines: three visual token pruning methods, VisPruner ~\mbox{~\citep{zhang2025vispruner}}, VisionZip~\citep{visionzip}, and SpecFlow~\citep{li2026specflow}, and two PTQ methods, QuaRot~\citep{ashkboos2024quarot} and FlatQuant~\citep{sun2025flatquant}. All experiments use NVIDIA RTX 4090 GPUs with 24\,GB of memory. End-to-end inference speed and peak GPU memory consumption are measured on a single GPU with a batch size of 1.

\begin{table}[t]
\centering
\caption{Main comparison of pruning, quantization and our joint method. E2E denotes the end-to-end speedup, and Mem. denotes the peak memory footprint. The best performance is marked in \textcolor{red!60!black}{red}.}
\label{tab:main_result}

\setlength{\tabcolsep}{2.5pt}
\renewcommand{\arraystretch}{1.08}
\scriptsize

\resizebox{\columnwidth}{!}{%
\begin{tabular}{c|cccccccc|ccc}
\toprule
\textbf{Methods}
& \textbf{MMMU}
& \textbf{VizWiz}
& \textbf{SQA}
& \textbf{MME}
& \textbf{GQA}
& \textbf{MMB$^{\mathrm{CN}}$}
& \textbf{POPE}
& \textbf{SEED}
& \textbf{Avg.}$\uparrow$
& \textbf{E2E}$\uparrow$
& \textbf{Mem.}$\downarrow$ \\
\midrule

\rowcolor{gray!18}
LLaVA-NeXT-7B
& \multicolumn{8}{c|}{
  \textit{Upper Bound, 2880 Tokens} \textbf{(100\%)}
}
& & & \\

\fpvalue{Vanilla}
& \fpvalue{36.7}
& \fpvalue{58.7}
& \fpvalue{69.4}
& \fpvalue{1725.1}
& \fpvalue{63.9}
& \fpvalue{58.8}
& \fpvalue{86.5}
& \fpvalue{69.1}
& \fpvalue{100\%}
& \fpvalue{1$\times$}
& \fpvalue{100\%} \\

\rowcolor{gray!18}
Pruning
& \multicolumn{8}{c|}{
  \textit{Retain 320 Tokens}
  \textcolor{green!50!black}{($\downarrow$ 88.9\%)}
}
& & & \\

VisPruner\venue{ICCV25}
& 35.3 & 56.2 & 66.8 & 1585.8
& 57.7 & 52.5 & 81.3 & 60.9
& 92.7\% & 1.63$\times$ & 96.9\% \\

VisionZip\venue{CVPR25}
& 35.9 & 57.5 & 66.4 & 1616.9
& 58.1 & 54.3 & 79.3 & 62.9
& 93.9\% & 1.95$\times$ & 98.6\% \\

SpecFlow\venue{ICML26}
& 36.6 & 57.8 & 67.7 & 1651.6
& 58.6 & 54.9 & 79.9 & 63.6
& 95.1\% & 1.62$\times$ & 98.8\% \\

\rowcolor{gray!18}
Quant
& \multicolumn{8}{c|}{\textit{W4A4}}
& & & \\

QuaRot\venue{NeurIPS24}
& 29.8 & 53.2 & 60.6 & 1596.9
& 61.1 & 37.3 & 83.7 & 65.7
& 87.8\% & 1.45$\times$ & 70.1\% \\

FlatQuant\venue{ICML25}
& 33.0 & 54.5 & 64.2 & 1713.2
& 62.1 & 54.8 & 84.9 & 68.4
& 95.3\% & 1.83$\times$ & 70.8\% \\

\rowcolor{gray!18}
Joint
& \multicolumn{8}{c|}{
  \textit{Retain 320 Tokens}
  \textcolor{green!50!black}{($\downarrow$ 88.9\%)}
  \textit{\& W4A4}
}
& & & \\

\rowcolor{blue!5}
P4Q\venue{Ours}
& 36.7 & 56.5 & 67.3 & 1706.2
& 59.2 & 51.7 & 87.8 & 62.3
& \textcolor{red!60!black}{95.5\%}
& \textcolor{red!60!black}{2.80$\times$}
& \textcolor{red!60!black}{51.3\%} \\

\midrule

\rowcolor{gray!18}
LLaVA-1.5-7B
& \multicolumn{8}{c|}{
  \textit{Upper Bound, 576 Tokens} \textbf{(100\%)}
}
& & & \\

\fpvalue{Vanilla}
& \fpvalue{34.8}
& \fpvalue{54.3}
& \fpvalue{66.3}
& \fpvalue{1650.7}
& \fpvalue{60.6}
& \fpvalue{54.2}
& \fpvalue{81.8}
& \fpvalue{63.8}
& \fpvalue{100\%}
& \fpvalue{1$\times$}
& \fpvalue{100\%} \\

\rowcolor{gray!18}
Pruning
& \multicolumn{8}{c|}{
  \textit{Retain 64 Tokens}
  \textcolor{green!50!black}{($\downarrow$ 88.9\%)}
}
& & & \\

VisPruner\venue{ICCV25}
& 33.3 & 54.2 & 67.0 & 1514.6
& 53.0 & 51.7 & 70.5 & 55.6
& 93.1\% & 1.23$\times$ & 97.9\% \\

VisionZip\venue{CVPR25}
& 35.6 & 54.5 & 67.1 & 1535.2
& 53.4 & 51.9 & 72.2 & 57.1
& 94.8\% & 1.46$\times$ & 97.8\% \\

SpecFlow\venue{ICML26}
& 36.7 & 54.2 & 66.9 & 1475.0
& 56.8 & 53.4 & 70.0 & 56.2
& 95.2\% & 1.35$\times$ & 98.6\% \\

\rowcolor{gray!18}
Quant
& \multicolumn{8}{c|}{\textit{W4A4}}
& & & \\

QuaRot\venue{NeurIPS24}
& 31.5 & 54.2 & 59.2 & 1274.1
& 56.2 & 33.5 & 69.9 & 59.2
& 86.2\% & 1.69$\times$ & 49.8\% \\

FlatQuant\venue{ICML25}
& 35.1 & 53.7 & 65.0 & 1644.0
& 54.4 & 47.5 & 82.6 & 57.9
& 95.8\% & 1.75$\times$ & 50.9\% \\

\rowcolor{gray!18}
Joint
& \multicolumn{8}{c|}{
  \textit{Retain 64 Tokens}
  \textcolor{green!50!black}{($\downarrow$ 88.9\%)}
  \textit{\& W4A4}
}
& & & \\

\rowcolor{blue!5}
P4Q\venue{Ours}
& 34.2 & 54.0 & 66.4 & 1585.3
& 55.2 & 48.3 & 83.7 & 58.8
& \textcolor{red!60!black}{96.1\%}
& \textcolor{red!60!black}{2.49$\times$}
& \textcolor{red!60!black}{42.0\%} \\

\bottomrule
\end{tabular}
}
\end{table}













\subsection{Main Results}
\label{sec:main results}

\begin{wraptable}{r}{0.6\linewidth}
\centering
\caption{P4Q Generalization on Qwen2.5-VL-7B.}
\label{tab:qwen25}
\setlength{\tabcolsep}{1 pt}
\renewcommand{\arraystretch}{1}
\scriptsize
\resizebox{0.6\columnwidth}{!}{%
\begin{tabular}{c|cccc|c}
\toprule
\textbf{Methods}
& \textbf{MMMU}
& \textbf{VizWiz}
& \textbf{SQA}
& \textbf{MMB$^{\mathrm{CN}}$}
& \textbf{Avg.}$\uparrow$ \\
\midrule

\rowcolor{gray!18}
Qwen2.5-VL-7B
& \multicolumn{4}{c|}{\textit{Upper Bound, 576 Tokens} \textbf{(100\%)}}
& \\

\fpvalue{Vanilla}
& \fpvalue{51.4}
& \fpvalue{71.2}
& \fpvalue{87.7}
& \fpvalue{81.6}
& \fpvalue{100\%} \\

\rowcolor{gray!18}

& \multicolumn{4}{c|}{\textit{Retain 192 Tokens} \textcolor{green!50!black}{($\downarrow$ 66.7\%)}}
& \\

SpecFlow\venue{ICML26}
& 49.0
& 69.1
& 82.7
& 77.8
& 95.5\% \\

\rowcolor{blue!5}
P4Q\venue{Ours}
& 50.7
& 69.4
& 83.5
& 78.4
& \textcolor{red!60!black}{96.8\%} \\

\rowcolor{gray!18}

& \multicolumn{4}{c|}{\textit{Retain 64 Tokens} \textcolor{green!50!black}{($\downarrow$ 88.9\%)}}
& \\

SpecFlow\venue{ICML26}
& 47.3
& 63.6
& 80.9
& 70.3
& 89.9\% \\

\rowcolor{blue!5}
P4Q\venue{Ours}
& 46.9
& 63.9
& 81.3
& 70.7
& \textcolor{red!60!black}{90.1\%} \\

\bottomrule
\end{tabular}
}
\end{wraptable}

Table~\ref{tab:main_result} compares P4Q with five representative baselines, across eight image-understanding benchmarks. The pruning-based configurations retain 11.1\% of the visual tokens, while the quantization-based configurations use W4A4 settings. It is worth emphasizing that the reported performance retention, end-to-end speedup, and peak GPU memory usage are all averaged across eight benchmarks. Overall, P4Q maintains comparable task performance on LLaVA-NeXT-7B while achieving an average $2.8\times$ end-to-end inference speedup and using only 51.3\% of the peak GPU memory required by the vanilla model. On LLaVA-1.5-7B, P4Q likewise maintains comparable performance while delivering an average $2.49\times$ end-to-end inference speedup and reducing peak GPU memory usage by 58\% relative to the vanilla model. Together, these results show that P4Q combines quantization-aware token selection with pruning-aware calibration to shorten visual-token sequences while using W4A4 weights and activations.  On the two evaluated LLaVA models, this co-design maintains comparable task performance while substantially reducing inference latency and peak GPU memory usage relative to the uncompressed baselines. Further details of the efficiency evaluation are provided in Appendix~\ref{sec:appendix_efficiency_evaluation}.










To further evaluate the generalizability of P4Q across different VLM architectures, we extend our experiments to Qwen2.5-VL-7B, whose vision backbone and multimodal projector differ substantially from those of the LLaVA family. As shown in Table~\ref{tab:qwen25}, P4Q achieves average performance comparable to SpecFlow under both visual-token pruning budgets. Notably, P4Q follows the same joint W4A4 quantization and pruning configuration as in Table~\ref{tab:main_result}, whereas SpecFlow is evaluated with visual token pruning alone. These results demonstrate that P4Q remains effective on VLM architectures beyond the LLaVA family. 

\subsection{Apple-to-Apple Comparison with Peer Work}

\begin{wraptable}{r}{0.6\linewidth}
\centering
\caption{Comparison on LLaVA-1.5-7B. QUOTA~\citep{li2026quota} reports performance retention (\%) from its original paper.}
\label{tab:quota_p4q_llava15}
\setlength{\tabcolsep}{2.5pt}
\renewcommand{\arraystretch}{1.08}
\scriptsize
\resizebox{0.6\columnwidth}{!}{%
\begin{tabular}{c|cccc|c}
\toprule
\textbf{Methods}
& \textbf{MME}
& \textbf{GQA}
& \textbf{POPE}
& \textbf{SEED}
& \textbf{Avg.}$\uparrow$ \\
\midrule

\rowcolor{gray!18}
& \multicolumn{4}{c|}{
  \textit{Upper Bound, 576 Tokens} \textbf{(100\%)}
}
& \\

\fpvalue{Vanilla}
& \fpvalue{1650.7}
& \fpvalue{60.6}
& \fpvalue{81.8}
& \fpvalue{63.8}
& \fpvalue{100\%} \\

\rowcolor{gray!18}
& \multicolumn{4}{c|}{
  \textit{Retain 20\% Tokens}
  \textcolor{green!50!black}{($\downarrow$ 80.0\%)}
  \textit{\& W4A4}
}
& \\

QUOTA\venue{arXiv26}
& 90.82\%
& 93.53\%
& 97.56\%
& 95.09\%
& 94.3\% \\

\rowcolor{gray!18}
& \multicolumn{4}{c|}{
  \textit{Retain 10\% Tokens}
  \textcolor{green!50!black}{($\downarrow$ 90.0\%)}
  \textit{\& W4A4}
}
& \\

QUOTA\venue{arXiv26}
& 90.32\%
& 90.36\%
& 96.01\%
& 93.02\%
& 92.4\% \\

\rowcolor{gray!18}
& \multicolumn{4}{c|}{
  \textit{Retain 64 Tokens}
  \textcolor{green!50!black}{($\downarrow$ 88.9\%)}
  \textit{\& W4A4}
}
& \\

\rowcolor{blue!5}
P4Q\venue{Ours}
& 1585.3
& 55.2
& 83.7
& 58.8
& 95.4\% \\

\bottomrule
\end{tabular}
}
\end{wraptable}

To the best of our knowledge, no formally published work has jointly combined quantization and pruning to accelerate VLMs. Nevertheless, several concurrent works have appeared on arXiv, which we also include in our comparisons. Results are listed in Table \ref{tab:quota_p4q_llava15}. The comparison shows that with merely 11\% token retention, our accuracy exceeds QUOTA’s result at 20\% token retention by 1.1 percentage points. When compared under the more similar setting (QUOTA with 10\% retained tokens), our method gains a 3 percentage point advantage.

\subsection{Ablation Study}
\label{sec:ablation study}

\begin{wraptable}{r}{0.6\linewidth}
    \centering
    \small
\caption{Ablation study of quantization-aware token selection and pruning-aware calibration. PC is pruned calibration. FQP is fake quant for pruning.}
    \label{tab:main_ablation}
    \setlength{\tabcolsep}{1pt}
    \renewcommand{\arraystretch}{1}
\resizebox{0.55\columnwidth}{!}{%
    \begin{tabular}{@{}l|cc|cc|c@{}}
        \toprule
        \multirow{3}{*}{\textbf{Row}}
        & \multicolumn{2}{c}{\textbf{Quantization}}
        & \multicolumn{2}{c|}{\textbf{Pruning}}
        & \multirow{3}{*}{\makecell[c]{\textbf{Avg. Accuracy}\\\textbf{Retention (\%)}}} \\
        \cmidrule(lr){2-3}
        \cmidrule(lr){4-5}
        & \textbf{W4A4}
        & \makecell[c]{\textbf{PC}}
        & \textbf{Pruning}
        & \makecell[c]{\textbf{FQP}}
        & \\
        \midrule
        \rowcolor{gray!18}1 \rowtag{Vanilla} & --           & --           & --           & --           & 100.0\% \\
        2           & $\checkmark$ & --           & $\checkmark$ & --           & 92.1\%  \\
        3           & $\checkmark$ & --           & $\checkmark$ & $\checkmark$ & 92.5\%  \\
        4           & $\checkmark$ & $\checkmark$ & $\checkmark$ & --           & 93.2\%  \\
        \addlinespace[1pt]
        \rowcolor{blue!5}5 \rowtag{Ours}
                    & $\checkmark$ & $\checkmark$ & $\checkmark$ & $\checkmark$ & \textcolor{red!60!black}{95.5\%} \\
        \bottomrule
    \end{tabular}
    }
\end{wraptable}

We evaluate the components of Quantization-Aware Pre-LLM Selection and Pruning-Aware Quantization Calibration on LLaVA-NeXT-7B. Table~\ref{tab:main_ablation} reports average accuracy retention across five settings. Row 1 is the vanilla FP16 model. W4A4 indicates post-training quantization, while Pruning indicates an 88.9\% visual-token pruning rate. When Pruned Calibration is disabled, quantization is calibrated on dense sequences. Fake Quant for Pruning applies the selector-only fake quantization described in Section~\ref{sec:pre_llm_pruning} to the projector features used for token scoring. When it is disabled, selection uses the native FP16 features. Row 5 is the full P4Q setting used in Table~\ref{tab:main_result}.

Specifically, Row 2 corresponds to the setting where the quantized model is first loaded, followed by token pruning. It achieves a performance 3.4 percentage points lower than the final joint optimization method. This verifies that the superiority of our approach does not merely stem from the simple combination of quantization and pruning, and demonstrates the necessity of jointly designing quantization and pruning strategies. Comparing the experimental results of Rows 3, 4 and 5 verifies the effectiveness of the quantization and pruning techniques proposed in this paper.

\section{Conclusion}
\label{sec:conclusion}
We rethink visual token pruning and quantization as a coupled compression problem and introduce P4Q, a practical co-design framework for efficient VLM inference. P4Q combines quantization-aware pre-LLM token selection with pruning-aware calibration to shorten visual-token sequences while using W4A4 weights and activations. Experiments across eight multimodal benchmarks on the two evaluated LLaVA models show comparable task performance under W4A4 quantization and an 88.9\% visual-token pruning rate, while controlled ablations on LLaVA-NeXT-7B support the benefit of combining the two design choices. Relative to the uncompressed baselines, P4Q achieves average end-to-end speedups of $2.8\times$ and $2.49\times$ on LLaVA-NeXT-7B and LLaVA-1.5-7B, respectively, while substantially reducing peak GPU memory usage.

\subsection*{AI Use Statement}
In this work, generative AI tools were used to assist with language editing and manuscript checks, including grammar, clarity, concision, presentation, and consistency. They were not used to generate the core research ideas or design the methodology or experiments. All AI-assisted edits and suggestions were carefully reviewed and approved by the authors. The authors take full responsibility for the final content of this work.

\bibliography{references}
\bibliographystyle{arxiv_preprint}

\newpage
\appendix
\section{Additional Details of Low-Bit Rank Shift Analysis}
\label{sec:appendix_rank_shift}

\paragraph{Experimental details.}
We collected visual token ranking statistics on LLaVA-NeXT-7B using 128 fixed samples from the ScienceQA training split with random seed 42. FP16 and W4A4 inference used identical images, questions, packed input sequences, attention masks, and positional indices to ensure token-wise correspondence between the two precision settings. All samples were processed with the complete visual-token sequence, and no visual-token pruning was applied during statistics collection.

Only the language-model decoder was quantized, while the vision tower and multimodal projector remained unchanged. The W4A4 setting applied symmetric round-to-nearest quantization with per-output-channel 4-bit weights and per-token 4-bit activations. Both weight and activation group sizes were set to $-1$, the activation clipping ratio was set to $1.0$, and no INT8 exception was applied to the down-projection layers. For each sample, the final non-special token of the user question was used as the text query, and only image patch tokens were included as candidate visual keys, excluding the row-newline tokens introduced by LLaVA-NeXT AnyRes packing. For each target decoder layer, visual-token importance was measured using the attention scores from the selected question token to the candidate visual patch tokens.

\paragraph{Sample-balanced W4A4 rank shift relative to FP16.} For each sample $n$ and decoder layer $\ell$, the visual tokens are ranked in descending order of their importance scores. The resulting rank is normalized by the number of valid visual patch tokens $N_n$:
\begin{equation}
r_{n,\ell,i}^{\tau}
=
\frac{
\operatorname{rank}_{\downarrow}^{\mathrm{avg}}
\left(s_{n,\ell,i}^{\tau}\right)-1
}{
N_n-1
},
\qquad
\Delta r_{n,\ell,i}
=
r_{n,\ell,i}^{\mathrm{W4A4}}
-
r_{n,\ell,i}^{\mathrm{FP16}},
\label{eq:appendix_normalized_rank_shift}
\end{equation}
where $\tau\in\{\mathrm{FP16},\mathrm{W4A4}\}$, and average ranks are assigned to tokens with tied importance scores. Accordingly, $r=0$ denotes the most important visual token and $r=1$ denotes the least important one. A positive $\Delta r$ indicates that W4A4 moves a token toward a less important position, whereas a negative $\Delta r$ indicates that the token is promoted toward a more important position. FP16 and W4A4 observations are aligned using each visual token's original packed index, ensuring that the displacement is computed between the same token under the two precision settings.

To construct Fig.~\ref{fig:p4q_motivation}(b), we divide the same-layer FP16 normalized rank into $B=20$ equal-width intervals, denoted by $\mathcal{B}_b$. For each sample, we first average the signed rank shifts of all visual tokens whose FP16 ranks fall within the same interval. The plotted value is then obtained by averaging these sample-level means over the 128 samples:
\begin{equation}
\overline{\Delta r}_{\ell,b}
=
\frac{1}{128}
\sum_{n=1}^{128}
\operatorname{Mean}_{\,i:\,r_{n,\ell,i}^{\mathrm{FP16}}\in\mathcal{B}_b}
\left(
\Delta r_{n,\ell,i}
\right).
\label{eq:appendix_rank_shift_profile}
\end{equation}
The horizontal coordinate is the center of the corresponding FP16-rank interval, while the vertical coordinate is $100\overline{\Delta r}_{\ell,b}$, expressed in normalized-rank percentage points. This sample-balanced aggregation gives every sample equal weight and prevents images containing more AnyRes visual tokens from dominating the curve. The shaded regions indicate 95\% confidence intervals obtained from 2,000 non-parametric bootstrap resamples over the 128 sample-level means.

Because positive and negative displacements can cancel, the signed profile alone does not measure the overall magnitude of ranking rearrangement. We therefore additionally report the sample-balanced mean absolute rank shift, denoted as PRAS:
\begin{equation}
\operatorname{PRAS}_{\ell}
=
\frac{1}{128}
\sum_{n=1}^{128}
\left(
\frac{1}{N_n}
\sum_{i=1}^{N_n}
\left|
\Delta r_{n,\ell,i}^{\mathrm{W4A4}\leftarrow\mathrm{FP16}}
\right|
\right).
\label{eq:appendix_pras}
\end{equation}
Here, $\ell$ denotes the decoder layer, $n$ indexes the 128 evaluation samples, $N_n$ is the number of valid visual patch tokens in sample $n$, and $i$ indexes an individual visual token. The quantity $\Delta r_{n,\ell,i}^{\mathrm{W4A4}\leftarrow\mathrm{FP16}}$ is the signed normalized-rank displacement of token $i$ under W4A4 relative to FP16, while the absolute-value operator removes its movement direction. The inner average measures the mean absolute rank shift within one sample, and the outer average assigns equal weight to all 128 samples. PRAS therefore measures the average magnitude of W4A4-induced rank rearrangement without cancellation between positive and negative displacements. For example, $\operatorname{PRAS}_{\ell}=0.216$ corresponds to an average absolute displacement of $21.6$ rank percentage points. The PRAS values shown in the legend are $7.7\%$, $9.4\%$, $12.4\%$, and $21.6\%$ for Layers 2, 4, 8, and 16, respectively.

\paragraph{Qualitative visualization of Top-$K$ support changes induced by rank shift.}
To complement the sample-balanced quantitative analysis, we directly visualize the selected visual-token masks for a specific sample from the ScienceQA training split. For each layer and precision, the retained set is constructed independently from the corresponding attention scores:
\begin{equation}
\mathcal{T}_{n,\ell}^{\tau}(K)
=
\operatorname{TopK}_{i}
\left(s_{n,\ell,i}^{\tau},K\right),
\qquad
\rho_{n,\ell}^{\mathrm{repl}}
=
1-
\frac{
\left|
\mathcal{T}_{n,\ell}^{\mathrm{FP16}}(K)
\cap
\mathcal{T}_{n,\ell}^{\mathrm{W4A4}}(K)
\right|
}{
K
}.
\label{eq:appendix_topk_replacement}
\end{equation}
Here, $\mathcal{T}_{n,\ell}^{\tau}(K)$ denotes the indices of the $K$ highest-scoring visual tokens under precision $\tau$, and $\rho_{n,\ell}^{\mathrm{repl}}$ measures the fraction of the FP16 Top-$K$ set that is replaced under W4A4.
\begin{figure}[!htpb]
    \centering
    \includegraphics[width=0.9\linewidth]{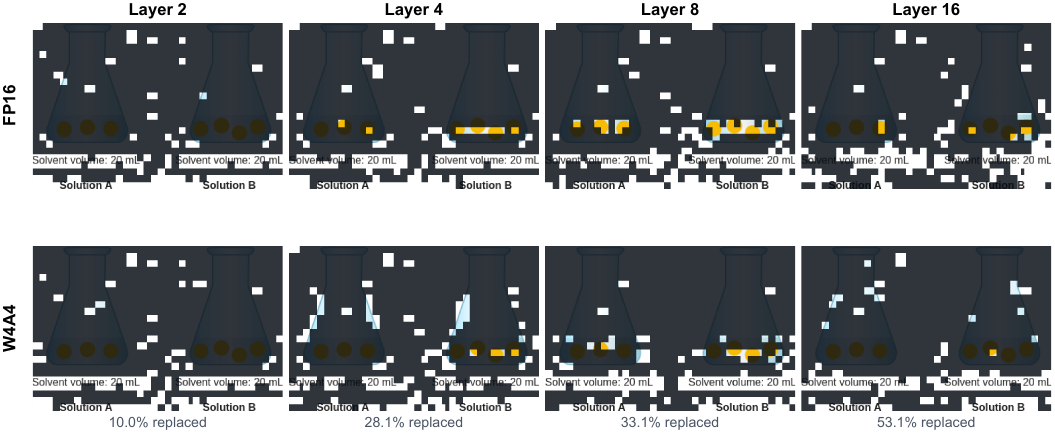}
    \caption{
    Qualitative visualization of the W4A4-induced visual-token selection changes for a specific sample. Each column compares the independently selected FP16 and W4A4 Top-$K$ sets at the same decoder layer. All settings
    retain $11.1\%$ visual patch tokens. Bright regions correspond to retained visual tokens, whereas dark regions correspond to masked tokens. The values below the W4A4 panels report the fraction of the
    FP16 Top-$K$ set replaced under W4A4: $10.0\%$, $28.1\%$, $33.1\%$, and $53.1\%$ at Layers 2, 4, 8, and 16, respectively.}
    \label{fig:appendix_sample002_rank_shift_masks}
\end{figure}

This visualization shows the spatial locations of the visual tokens retained under FP16 at representative decoder layers and compares them with the tokens retained by the same selection strategy at the corresponding layers under W4A4. This single-sample comparison provides an intuitive qualitative illustration of how W4A4-induced rank shift changes the retained token set, whereas Fig.~\ref{fig:p4q_motivation}(b) presents the corresponding quantitative analysis over all 128 samples.

\newpage
\section{Additional Experimental Details}

This section provides additional implementation details for the experiments reported in the main manuscript, including the experimental configurations, calibration and evaluation settings, hyperparameter choices, and statistical procedures. These details complement the descriptions in the main text and facilitate the interpretation and reproduction of our experiments. By documenting the settings underlying the reported results, we aim to provide a more transparent account of the empirical evidence supporting our main findings and conclusions.

\subsection{Experimental Settings}
\label{sec:appendix_experimental_settings}
All experiments were conducted on a server equipped with NVIDIA GeForce RTX 4090 GPUs, each with 24\,GB of memory. The implementation used Python 3.10.20, PyTorch 2.6.0 and CUDA 11.8. We evaluate P4Q on a diverse suite of eight established image-understanding benchmarks, including MMMU~\citep{yue2024mmmu}, VizWiz~\citep{gurari2018vizwiz}, ScienceQA\_TEST (SQA)~\citep{lu2022learnsqa}, MME~\citep{fu2025mme}, GQA~\citep{hudson2019gqa}, MMBench\_CN (MMB$^{\mathrm{CN}}$)~\citep{liu2024mmbench}, POPE~\citep{li2023evaluatingpope}, and SEEDBench\_IMG (SEED)~\citep{li2023seed}. All benchmarks are evaluated using VLMEvalKit~\citep{duan2024vlmevalkit}, following their official evaluation protocols and default settings.

\subsection{Sensitivity to Token-budget Allocation}
\label{sec:appendix_pruning}
\begin{table}[!htbp]
\centering
\caption{Sensitivity analysis of the token-budget allocation hyperparameters on LLaVA-1.5-7B at an 88.9\% visual-token pruning ratio. $\alpha$, $\beta$, and $\gamma$ denote the proportions allocated to text-guided anchors, visual-saliency anchors, and diversity completion.}
\label{tab:hyperparameter_ablation}
\setlength{\tabcolsep}{4pt}
\resizebox{0.65\columnwidth}{!}{%
\begin{tabular}{@{}l|ccc|ccc|c@{}}
\toprule
\multirow{3}{*}{\textbf{Row}}
& \multicolumn{3}{c|}{\textbf{Token-Budget}}
& \multicolumn{3}{c|}{\textbf{Benchmark}}
& \multirow{3}{*}{\makecell[c]{\textbf{Avg. Accuracy}\\\textbf{Retention (\%)}}} \\
\cmidrule(lr){2-4}
\cmidrule(lr){5-7}
& $\boldsymbol{\alpha}$
& $\boldsymbol{\beta}$
& $\boldsymbol{\gamma}$
& MMMU
& SQA
& MME
& \\
\midrule

\rowcolor{gray!18}
Vanilla
& -- & -- & --
& 34.8 & 66.3 & 1650.7
& 100.0\% \\

1
& 0.42 & 0.28 & 0.30
& 34.3 & 65.5 & 1600.8
& 98.1\% \\

2
& 0.20 & 0.30 & 0.50
& 34.9 & 66.0 & 1564.5
& 98.2\% \\

3
& 0.40 & 0.10 & 0.50
& 34.7 & 65.7 & 1574.7
& 98.1\% \\

4
& 0.12 & 0.18 & 0.70
& 33.3 & 66.3 & 1586.1
& 97.3\% \\

5
& 0.18 & 0.12 & 0.70
& 34.0 & 65.9 & 1572.9
& 97.5\% \\

6
& 0.24 & 0.06 & 0.70
& 34.0 & 66.1 & 1545.2
& 97.0\% \\

\addlinespace[1pt]
\rowcolor{blue!5}
7 \rowtag{Ours}
& 0.30 & 0.20 & 0.50
& 34.2 & 66.4 & 1585.3
& 98.2\% \\

\bottomrule
\end{tabular}%
}
\end{table}
Table~\ref{tab:hyperparameter_ablation} evaluates the sensitivity of P4Q to the allocation ratios $\alpha$, $\beta$, and $\gamma$ while keeping the total visual-token budget and the 88.9\% pruning ratio fixed. Across all tested allocations, the average performance retention remains between $97.0\%$ and $98.2\%$, corresponding to a maximum variation of only $1.2$ percentage points. The variations on the individual benchmarks are also limited: MMMU ranges from $33.3$ to $34.9$, ScienceQA from $65.5$ to $66.4$, and MME from $1545.2$ to $1600.8$. In particular, when the diversity budget is fixed at $\gamma=0.5$, substantially redistributing the remaining budget between Query Relevance and Visual Saliency changes the average performance retention by at most $0.1$ percentage points. Allocating a larger proportion to Visual Diversity ($\gamma=0.7$) results in a small reduction to $97.0$--$97.5\%$, suggesting that excessive diversity completion may reduce the budget available for query-relevant and visually salient evidence. We therefore adopt $\alpha=0.3$, $\beta=0.2$, and $\gamma=0.5$ as a balanced default. This setting achieves $98.2\%$ average performance retention, tying the best result among the evaluated allocations. Overall, the results show that the three budget ratios can cause modest benchmark-level fluctuations but that P4Q remains robust to reasonable variations in their values.

\subsection{PTQ Calibration Implementation Details}
\label{sec:appendix_calibration_implementation}
For the W4A4 setting, we quantized the language-model decoder to 4-bit weights and 4-bit activations while retaining the key and value caches in FP16; the vision tower and multimodal projector remained in full precision. Weight quantization was symmetric and per output channel, activation quantization was symmetric and per token, and no additional group-wise subdivision was used. We sampled 128 fixed image-question pairs from the ScienceQA training split with random seed 42 as the calibration set. To align calibration with the deployment-time inference path, the same P4Q pruning strategy and model-specific visual-token retention ratio used during inference were applied before capturing the decoder calibration inputs. Calibration was then performed sequentially over the decoder blocks. For each block, its full-precision outputs on the pruned calibration sequences were used as reconstruction targets, and the corresponding quantized block was optimized by minimizing the mean-squared reconstruction error. The learnable calibration parameters included factorized channel transformations, per-channel diagonal scales, and learnable weight and activation clipping factors; the original model parameters were frozen. The diagonal scales were initialized from the observed activation ranges and weight ranges using an interpolation exponent of $0.3$. Each decoder block was calibrated for 15 epochs with an effective batch size of 1 using AdamW. The learning rate was set to $5\times10^{-3}$ for the transformation matrices and diagonal scales and $5\times10^{-2}$ for the clipping parameters, followed by cosine annealing to $5\times10^{-6}$. After calibration, the learned transformations and diagonal scales were folded into the decoder weights, and the resulting weights were quantized using round-to-nearest quantization for W4A4 inference.

\subsection{Efficiency Evaluation Details}
\label{sec:appendix_efficiency_evaluation}

This section provides additional details on the computational efficiency improvements reported in Table~\ref{tab:main_result}. Specifically, E2E denotes the average per-sample end-to-end latency measured across the eight evaluation datasets with a batch size of 1. The latency measurement covers both data loading and the complete VLM inference pipeline. All experiments were conducted on a single NVIDIA GeForce RTX 4090 GPU with 24 GB of memory. Table~\ref{tab:efficiency_details} presents the dataset-level measurements underlying the average efficiency improvements reported in Table~\ref{tab:main_result}. Notably, for LLaVA-NeXT-7B, the end-to-end inference speedups on MMMU, MME, and POPE all exceeded \textcolor{red!60!black}{\textbf{3$\times$}}, with the largest speedup of \textcolor{red!60!black}{\textbf{3.26$\times$}} observed on MME.

\begin{table*}[!htbp]
\centering
\caption{End-to-end inference latency and peak GPU memory of P4Q on LLaVA-NeXT-7B and LLaVA-1.5-7B. Latency is reported as the mean processing time per sample in milliseconds, and peak memory is reported in GB. For latency, the Avg. column reports the average dataset-wise speedup over Vanilla. For memory, it reports the average memory consumption relative to Vanilla. Lower latency and memory consumption are better.}
\label{tab:efficiency_details}
\setlength{\tabcolsep}{4pt}
\resizebox{\textwidth}{!}{
\begin{tabular}{@{}ll|cccccccc|c@{}}
\toprule
\textbf{Metric}
& \textbf{Method}
& \textbf{MMMU}
& \textbf{VizWiz}
& \textbf{SQA}
& \textbf{MME}
& \textbf{GQA}
& \textbf{MMB$^{\mathrm{CN}}$}
& \textbf{POPE}
& \textbf{SEED}
& \textbf{Avg.} \\
\midrule

\rowcolor{gray!18}
\multicolumn{11}{c}{\textit{LLaVA-NeXT-7B}} \\

E2E Latency
& Vanilla
& 2144.33 & 2031.77 & 635.79 & 886.13
& 378.12 & 1044.38 & 552.55 & 367.08
& 1.00$\times$ \\

\rowcolor{blue!5}
(ms/sample)
& P4Q
& 707.36 & 806.95 & 246.78 & 271.57
& 165.86 & 367.40 & 179.10 & 130.63
& 2.80$\times$ \\

\cmidrule(lr){2-10}

Peak Memory
& Vanilla
& 20.31 & 14.88 & 14.95 & 14.90
& 14.90 & 15.14 & 14.89 & 14.95
& 100.0\% \\

\rowcolor{blue!5}
(GB)
& P4Q
& 14.64 & 7.22 & 7.08 & 7.02
& 6.62 & 8.01 & 7.29 & 7.32
& 51.3\% \\

\midrule

\rowcolor{gray!18}
\multicolumn{11}{c}{\textit{LLaVA-1.5-7B}}\\

E2E Latency
& Vanilla
& 600.48 & 1001.80 & 184.44 & 411.05
& 361.51 & 367.36 & 354.53 & 297.55
& 1.00$\times$ \\

\rowcolor{blue!5}
(ms/sample)
& P4Q
& 230.92 & 366.19 & 111.12 & 178.87
& 125.90 & 165.15 & 114.86 & 121.77
& 2.49$\times$ \\

\cmidrule(lr){2-10}

Peak Memory
& Vanilla
& 15.00 & 13.56 & 13.82 & 13.53
& 13.53 & 14.10 & 13.52 & 13.58
& 100.0\% \\

\rowcolor{blue!5}
(GB) 
& P4Q
& 7.31 & 5.34 & 5.67 & 5.48
& 4.66 & 6.61 & 5.73 & 5.77
& 42.0\% \\

\bottomrule
\end{tabular}
}
\end{table*}

We further investigate and report the stage-wise acceleration underlying the end-to-end inference gains. Specifically, on LLaVA-NeXT-7B, we measure the average per-sample prefill latency across eight benchmarks. The reported prefill latency covers the language-model prefill computation from cache readiness to prefill completion. As shown in Fig.~\ref{fig:prefill_speedup}, P4Q consistently achieves substantial prefill acceleration across all eight benchmarks, with an average speedup of \textcolor{red!60!black}{\textbf{3.33$\times$}} over FP16. In particular, the largest improvement is observed on VizWiz, where P4Q achieves a prefill speedup of \textcolor{red!60!black}{\textbf{4.58$\times$}}.

\clearpage

\begingroup
\makeatletter
\setlength{\@fptop}{0pt}
\setlength{\@fpbot}{0pt plus 1fil}
\makeatother
\begin{figure}[p]
    \centering
    \includegraphics[width=0.75\columnwidth]{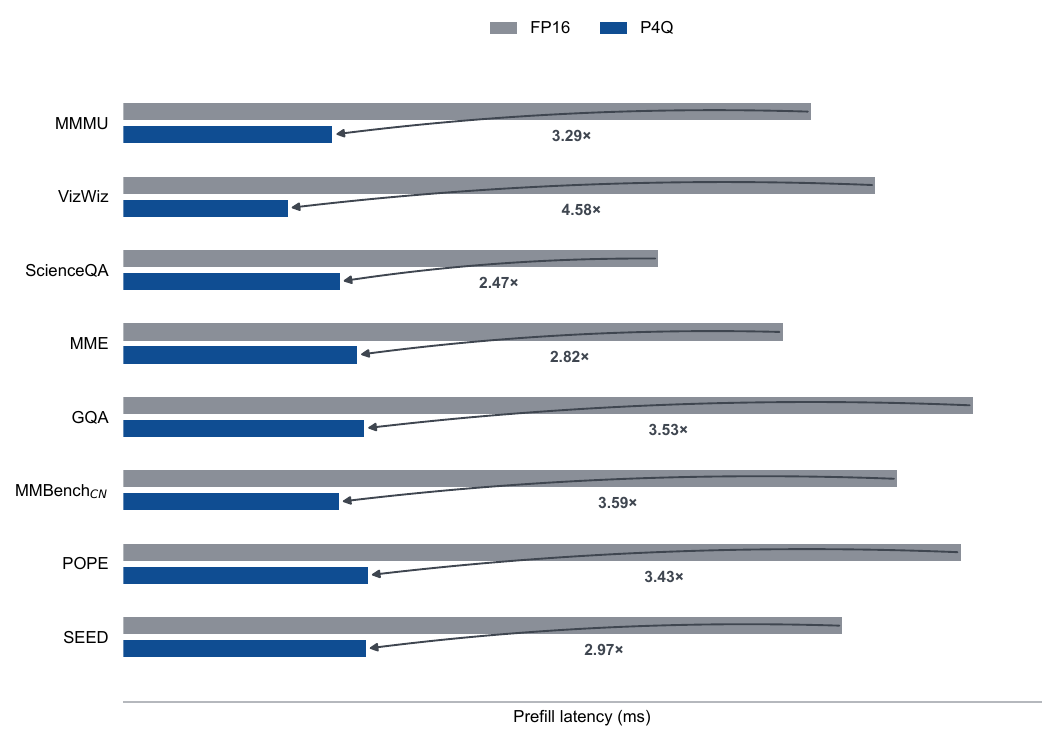}
    \caption{Per-sample prefill latency and speedup of P4Q over FP16 on eight benchmarks using LLaVA-NeXT-7B. Each pair of horizontal bars compares the average prefill latency of FP16 and P4Q, while the annotated values indicate the corresponding speedup. P4Q achieves an average prefill speedup of $3.33\times$.}
    \label{fig:prefill_speedup}
\end{figure}

\clearpage
\endgroup

\end{document}